# Equipment Failure Analysis for Oil and Gas Industry with an Ensemble Predictive Model

Chen ZhiYuan[1], Olugbenro. O. Selere[2] and Nicholas Lu Chee Seng[3]

[1]School of Computer Science, University of Nottingham Malaysia Campus, Semenyih, Malaysia
[2]School of Computer Science, University of Nottingham Malaysia Campus, Semenyih, Malaysia
[3]Tentacle Technologies, Kuala Lumpur, Malaysia
Zhiyuan.chen@nottingham.edu.my, Khcx4oor@nottingham.edu.my, nicklu@tentacletech.com

*Abstract*

This paper aims at improving the classification accuracy of a Support Vector Machine (SVM) classifier with Sequential Minimal Optimization (SMO) training algorithm in order to properly classify failure and normal instances from oil and gas equipment data. Recent applications of failure analysis have made use of the SVM technique without implementing SMO training algorithm, while in our study we show that the proposed solution can perform much better when using the SMO training algorithm. Furthermore, we implement the ensemble approach, which is a hybrid rule based and neural network classifier to improve the performance of the SVM classifier (with SMO training algorithm). The optimization study is as a result of the underperformance of the classifier when dealing with imbalanced dataset. The selected best performing classifiers are combined together with SVM classifier (with SMO training algorithm) by using the stacking ensemble method which is to create an efficient ensemble predictive model that can handle the issue of imbalanced data. The classification performance of this predictive model is considerably better than the SVM with and without SMO training algorithm and many other conventional classifiers.

## 1. INTRODUCTION

In the oil and gas industry it has become of primary importance that assets are properly used and maintained as the cost of failure for these assets is rather expensive. Failure or incidents are events that occur in rarity but when they do the effect they leave can be very harmful. Proper understanding of how the use of an asset can lead to failure is vital, as Oil and Gas industries need to address Health, Safety and Environment issues. Equipment monitoring at oil and gas plants can be a very difficult task, because of the varying conditions plant operators have to face as regards to the specific equipment.

The purpose of this work is to develop a solution using machine learning for data-driven failure analysis. The solution is to properly classify failure and normal events from oil and gas data. The main classifier used towards implementing this solution is the Sequential Minimal Optimizer (SMO) which is used to train a support vector machine. This classifier was chosen because of the faster approach taken towards the quadratic programming used by the original SVM created by Vapnik (1992), and also because the dataset used presented a class imbalance problem.

This class imbalance was a result of having more normal class labels in the data than the failure instances, and as a result of this, the SVM training classifier (SMO) needed to be optimized to handle such data. This optimization approach led to the development of the ensemble model using the ensemble learning approach called stacking. The ensemble approach was taken because of the need to ensure that the classification algorithm selected did not attempt to over-fit the data during training because of the imbalance. Through the stacking approach, the SVM training classifier (SMO) would learn from the prediction approach used by the base classifiers to better predict the normal and failure events in the resulting dataset

Three classifiers were selected for the stacking approach with the SMO; the random forest decision tree algorithm, the partial-decision trees (PART) rule-based algorithm, and the multilayer perceptron which is a neural network algorithm. These algorithms were selected because of their predictive performance on the imbalanced dataset after measures of treatment such as; oversampling using the synthetic minority oversampling technique (SMOTE) and the under-sampling technique had been applied to the dataset.

Through the stacking approach, the SVM training classifier (SMO) would learn from the prediction approach used by the base classifiers to better predict the normal and failure events in the resulting dataset. The resulting model was a better performer than the SMO on its own after conducting tests on the dataset without any sampling or cost sensitive learning technique applied to the dataset. This paper is divided into 6 sections. Section I of this paper presents a review of literature as regards to predictive analytics and its use in the oil and gas industry and recent applications of failure analysis. Section II presents the methodology towards the achievement of the project aim and goals. Section III presents the experiment setup, with all the necessary steps taken towards the creation of the ensemble model. Section IV presents the results from the experiments conducted. Section V gives a discussion about the results conducted from the experiments and the final section gives a summary of our findings and any future implementations that can be done towards solving the problem.
.

## 2. LITERATURE REVIEW

In this section of the paper, an overview of predictive analytic, its implementations and the recent industry applications in the oil and gas sector have been studied.

### 2.1 Predictive Analytics

Predictive analytics is a branch of machine learning. The notion behind predictive analytics involves making future predictions about events after conducting techniques encompassing machine learning and data mining on current and historical data[1]. The core of predictive analytics relies on capturing the possible relationships existing between variables and the predictive variables from past occurrences, to predict future outcomes[2]. Predictive modelling draws from statistics and optimization techniques to extract accurate information from large volumes of data. From such modelling can oil and gas personnel produce interpretable information that can be used to understand the implications of events, enabling them to take action based on these implications[3].

In the oil and gas industry the system in place for conducting asset maintenance was the enterprise asset management system (EAM). Predictive analytics builds on this system by combining real-time data from sensor placed in equipments with historical data to predict potential asset failures. This then enables the move from a reactive mode of maintenance whereby equipment has to fail before fixed, to proactive maintenance whereby equipment can be maintained before failure[3].

### 2.2 Applications of Failure Analysis

A review of literature revealed some recent application of failure analysis in the oil and gas industry. The work proposed by Saybani et al. (2011) shows how a machine learning technique called anomaly detection can be used in the detecting failure in sensor data and also to predict the next occurrence of a sensor failure[4]. They highlight the fact that currently, human experts try to detect sensor failures or anomalies; they do this by watching and studying the data during production process, but they have limits to how much they can observe properly. To cluster sensor data, a fuzzy-based predictor model was generated automatically using subtractive fuzzy clustering method. Derived time series, a new kind of time series is introduced and proposed. Furthermore, this paper shows two prediction models namely: auto regressive integrated moving average and autoregressive tree models which are used for predicting the next occurrence of sensor failure. As Singh (2006) mentions, "*reducing equipment downtime, increasing reliability and availability of the equipments are considered as the most important strategical objectives, which can optimize the life cycle of the equipment*". He considers costs associated with manufacturing design as fixed and predetermined, and therefore he suggests, in order to be competitive in the open market, the users have no other choice than optimizing life cycle of engines during their operation and maintenance. In the context of this paper, "engines" here corresponds to the equipment used in a refinery[5]. Experiments were conducted using two algorithms: ARIMA a predictive modelling technique used in SPSS and Forecast a function in Microsoft Excel with a connection to SQL Server 2008 to check all sensor values in conducting analysis on time related data.

The focus of the work proposed by Patri et al (2014) is to demonstrate how a time series analysis approach can be applied to failure detection and failure prediction from the streams of sensor data[6]. This method involves identifying *shapelets* which are short instances that are particularly distinct in the streams of sensor data. The streams of data obtained are from electrical submersible pumps, each instrumented with sensors that continually measure electrical properties of the pump. The *shapelets* approach is particularly appropriate for large oil and gas enterprise datasets because the algorithm does not access of the historical data. This greatly reduces the amount of data that needs to be stored for analysis. Moreover, unlike model-based approaches, shapelet-based analysis does not make any assumptions about the underlying nature of the data, making it practical for applications where a detailed physical model of the pump is not available. Examples of such are conducting experiments in a simulated laboratory[7]. The final results from failure prediction and detection were then compared with other classifiers, and that showed that the time-series *shapelets* approach was a better performer in failure detection than the other classifiers but in failure prediction, outperformed by the SVM (rbf), AdaBoost, J48 and RF(random forest) classifiers. The method is able to detect failures only from intake pressure measurements with a precision of 89% and 100% recall. The accuracy of predicting failures is lower with a precision of 78% at 78% recall. Towards better prediction and detection, multiples measurement attributes can be considered.

Also in the work proposed by Raghevendra et. Al (2011), they proposed a failure detection approach which can be applied to rod pump wells. The proposed approach involves a system built for recognizing pumps which are either failing, failed, or those still in normal working condition. The proposed system works in a two-step process where a machine learning algorithm using a method called boosting learns from several Bayesian Network models, and then combines these models to build a better efficient model. The newly formed model is then applied to all the wells in the field as opposed to methods of before which aimed to build a model per well. This model performs rather well, as it is able to detect failures with an accuracy level higher than 90%[8,9].

## 3. METHODOLOGY

This section of this paper presents the various methods and approaches that were used towards the development of the proposed solution.

### 3.1 Unsupervised Learning

Supervised learning entails using a function to learn a mapping between a set of input variables and an output variable, and applying this mapping function to predict the outputs for unseen data[10]. Unlike supervised learning, unsupervised learning involves drawing inference from unlabeled data. The method of unsupervised learning under focus is clustering analysis. Cluster analysis involves the creation of different categories containing objects of similar characteristics. In other words cluster analysis aims at placing data objects into different groups depending on the degree of association between the data objects in order to determine if they belong to that group or not. Figure 1 below shows a typical clustered dataset[11].

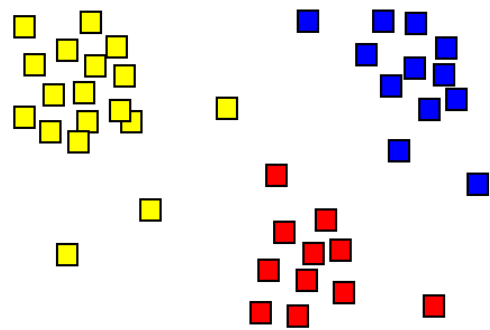

**Figure 1.** Three clusters to represent a dataset

A typical application of this analysis is the determining the number of states in a country with high rate of crime. Based on the available factors for each state such as rate of unemployment, number of graduates, income and other such factors cluster analysis can then be done to group the states into clusters. From this clusters inference can then be drawn on which particular factor influences the clusters created.

Various algorithms exist for conducting clustering analysis such as the popular *k*-means algorithm, but the clustering algorithm considered is the expectation maximization (EM). The EM algorithm approaches the clustering technique in a different way from the k-means algorithm by estimating the cluster to which an observation belongs by computing probabilities based on one or more probability distributions. The overall aim of this algorithm is to maximize the cluster an observation would belong to given the number of clusters to assign observations to. Because of this uniqueness from the k-means, the EM algorithm can also be applied to both continuous and categorical variables. Figure 2 below shows the clusters generated by the expectation maximization algorithm for Gaussian-distributed data[12].

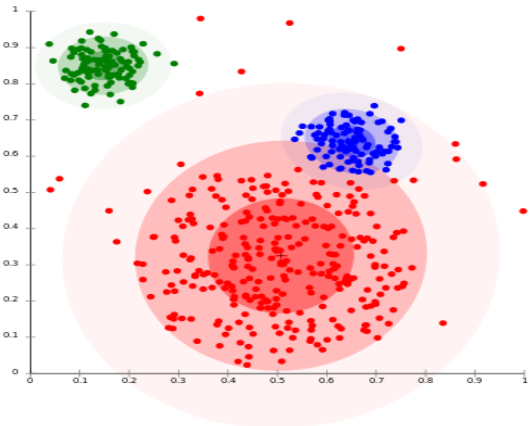

**Figure 2.** Gaussian data created by EM

The EM algorithm requires us to iterate through the following two steps:
- **The Expectation Step**: Using the current best guess for the parameters of the data model, we construct an expression for the log-likelihood for all data, observed and unobserved, and, then, marginalize the expression with respect to the unobserved data. This expression will be shown to depend on both the current best guess for the model parameters and the model parameters treated as variables.
- **The Maximization Step**: Given the expression resulting from the previous step, for the next guess we choose those values for the model parameters that maximize the expectation expression. These constitute our best new guess for the model parameters.

The output of the Expectation Step codifies our expectation with regard to what model parameters are most consistent with the data actually observed and with the current guess for the parameters provided we maximize the expression yielded by this step. We stop iterating through the two steps when any further change in the log-likelihood of the observed data falls below some small threshold[13].

### 3.2 Sequential Minimal Optimization

The sequential minimal optimization algorithm created by (Platt, 1998) is a simple algorithm that can quickly solve the SVM Quadratic Programming (QP) problem without any extra matrix storage and without using numerical QP optimization steps at all. SMO decomposes the overall QP problem (see figure 5) into QP sub-problems, using Osuna's theorem to ensure convergence[14].

To solve the QP problem below

$$\max_{\boldsymbol{\alpha}} W(\boldsymbol{\alpha}) = \sum_{i=1}^{\ell} \alpha_i - \frac{1}{2} \sum_{i=1}^{\ell} \sum_{j=1}^{\ell} y_i y_j k(\vec{x}_i, \vec{x}_j) \alpha_i \alpha_j,$$
$$0 \leq \alpha_i \leq C, \quad \forall i,$$
$$\sum_{i=1}^{\ell} y_i \alpha_i = 0.$$

**Figure 3.** SVM QP Problem

the SMO aims to satisfy the Karush-Kuhn-Tucker (KKT) conditions[15]. Platt when describing KKT says "*a point is an optimal point of (see figure 4) if and only if the Karush-Kuhn-Tucker (KKT) conditions are fulfilled and $Q_{ij} = y_i y_j k(\vec{x}_i, \vec{x}_j)$ is positive semi-definite. Such a point may be a non-unique and non-isolated optimum*"[16]. These KKT conditions which are particularly simple, allow for the QP to be solved when for all *i*:

$$\alpha_i = 0 \Rightarrow y_i f(\vec{x}_i) \geq 1,$$
$$0 < \alpha_i < C \Rightarrow y_i f(\vec{x}_i) = 1,$$
$$\alpha_i = C \Rightarrow y_i f(\vec{x}_i) \leq 1.$$

**Figure 4.** KKT Conditions

Unlike the methods for solving the SVM QP problem, SMO looks to solve the smallest possible optimization problem at every step. In order to do this the SMO chooses two Lagrange multiplier which it jointly optimizes, after which it finds the optimal values for these multipliers, and updates the SVM with the new values obtained from the optimization step. The reason for selecting two Lagrange multipliers at every step is because in the standard SVM QP problem, the smallest possible optimization problem involves two Lagrange multipliers, and these multipliers must obey a linear equality constraint[16]. (See figure 6).

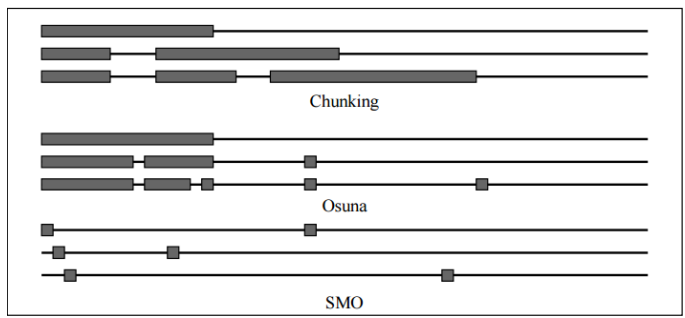

**Figure 5.** Comparison of SVM learning methods

For each method, three steps are illustrated. The horizontal thin line at every step represents the training set, while the thick boxes represent the Lagrange multipliers being optimized at that step. A given group of three lines corresponds to three training iterations, with the first iteration at the top

What gives the SMO the advantage over other SVM training methods is the analytical way in which it solves the tow Lagrange multipliers. This analytic method makes the SMO avoid the numerical QP optimization step. This is because the SMO solves as many optimization sub-problems which makes it solve the overall QP problem very quickly[16]. Also because the SMO doesn't use matrix algorithms, it is less likely to be susceptible to numerical precision problems[16].

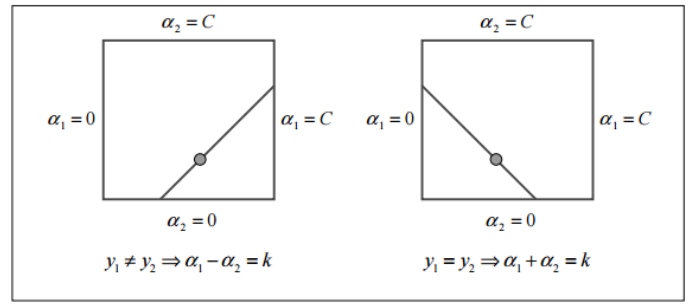

**Figure 6.** Two Lagrange multipliers

The two Lagrange multipliers as shown in figure 6 above must fulfil all of the constraints of the full problem[16]. The inequality constraints cause the Lagrange multipliers to lie in the box. The linear equality constraint causes them to lie on a diagonal line. Therefore, one step of SMO must find an optimum of the objective function on a diagonal line segment[16].

### 3.3 Ensemble Learning

Ensemble learning is the process that combining classifiers to solve a computational intelligence problem. It is only used in order to improve the predictive performance of a model, or in other instances to reduce the likelihood of having selected a wrong model[17].

There are three common methods of ensemble learning which are; boosting, bagging and stacking. Bagging is one of the most intuitive and simplest to implement, and tends to perform considerably well[18]. In boosting, resampling is strategically geared to provide the most informative training data for each consecutive classifier.

Stacking is the method of involving an ensemble of classifiers and it's outputs are used as inputs to a second-level meta-classifier to learn the mapping between the ensemble outputs and the actual correct classes. In essence, when a particular classifier incorrectly misclassifies instances from a particular region as a result of incorrectly learning the region of the feature space, a second-level meta-classifiers can then learn from this behavior and that of other classifiers to correct such improper training[17]. The figure below illustrates the stacked generalization approach, where classifiers $C_1,...,C_T$ are trained using training parameters $\theta_1$ through $\theta_T$ to output hypotheses $h_1$ through $h_T$. The outputs of these classifiers and the corresponding true classes are then used as input/output training pairs for the

second level classifier, $C_{T+1}$. Figure 7 below[17] shows this

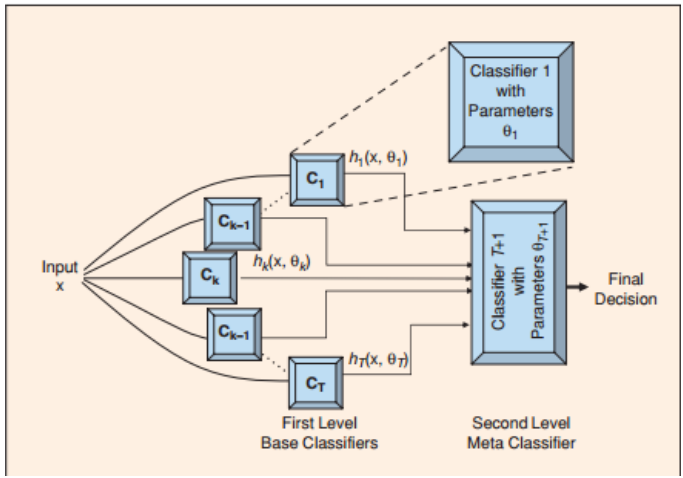

**Figure 7.** Stacking technique

### 3.4 Class Imbalance

An imbalanced problem occurs when one of the two classes in a dataset contains more samples than the other class. The result of this makes the class with the more samples, the majority class while the class with the least amount of sample, the minority class[19]. This then creates a situation where classifiers tend to show bias towards the majority class during prediction, and end up predicting poorly on the minority class[19].

Numerous techniques exist for solving class imbalance related problems, and these techniques have been further broken down into three categories: the algorithmic, data-preprocessing, and the feature selection approach. The data preprocessing approach is the approach used towards the proposed solution. With the data preprocessing technique sampling is applied on the data in which either new samples are added or existing samples are removed in order to reduce the class imbalance in the dataset.

The addition of new samples to the data is called oversampling and the technique selected for applying this approach is the synthetic minority oversampling technique (SMOTE). This sampling technique created by Chawla et al 2002, is to create new samples for the minority class, thereby causing the number of samples in the dataset to increase[20]. The minority class is over-sampled by taking each minority class sample and introducing synthetic examples along the line segments joining any/all of the k minority class nearest neighbors[20]. Figure 8 shows the application of oversampling to a dataset[19].

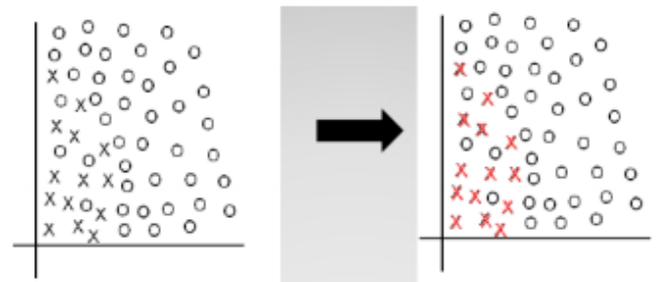

**Figure 8.** Oversampling of the minority class

The second sampling technique is the under-sampling technique. With this technique the majority class is under-sampled by randomly removing samples until there is an equal balance between the number of minority samples and majority samples[20]. This allows for the learner not to show
bias when conducting predictions on the dataset. One drawback of this technique is that because it uses a random under-sampling approach, useful data can be discarded as a result of randomly removing majority class samples, which would have been of great importance to the classifiers[21]. Figure 9 below shows the application of under-sampling to data[19].

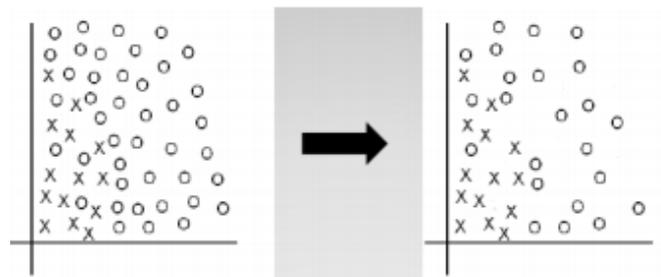

**Figure 9.** Undersampling of the majority class

Another technique of dealing with the class imbalance problem is to use cost sensitive learning. The goal of this type of learning is to minimize the total cost in properly predicting the class for a data point. To properly minimize the total cost, a cost matrix is inserted to the training algorithm during training, and for the weights specified in the matrix, they become the cost of misclassifying an instance during classification and prediction. This approach tends to make the classifier a much better performer when making predictions.

## 4. EXPERIMENT SETUP

This section introduces the experiment setup and the necessary steps taken to perform proper analysis of the dataset, towards the proposed solution.

### 4.1 Approach

Experiments were performed in two stages. The first sets of experiments are conducted to compare a set of classifiers with the SVM training classifier (SMO) with no sampling, oversampling, under-sampling, and cost sensitive classification applied to the dataset. From the experiments, the best performing classifiers are selected by their performance measures. The second set of experiments is to compare the ensemble classifier of the SMO combined with the best performing classifiers with the initial set of classifiers used in the first set of experiments to see the performance on the imbalanced dataset when no form of treatment has been applied.

### 4.2 Dataset Description

The dataset used contained 9 attributes and 875,000 instances. The data was collected from an online source for a study conducted on oil and gas pipes to determine the hazard levels. The dataset was unlabeled prompting the need for a machine learning approach to give class labels to the data before conducting predictions. Also the dataset comprised of numeric attributes and the class label as the nominal attribute. There was no case of missing data in the dataset and the dataset contained continuous variables. The data recordings were taken for 15 days starting from 16/12/2014 to 31/12/2014, with features such as operating temperature (°C), working pressure (psi), flow rate (cc/min) etc. The sample dataset is shown below:

**Figure 10.** Sample dataset

### 4.3 Classifiers Selected

Classifiers were selected based on the literature review and the reduction of variance and bias when learning from the training set. Therefore Multilayer perceptron (MLP), Naïve Bayes (NB), PART, LibSVM, J48 and the RandomForest (RF) were selected. These classifiers were then used for comparison with the SVM training classifier (SMO) to measure the performance.

### 4.4 Data Preprocessing

As the dataset was unlabeled an approach was used to assign class labels to the data. This approach was an unsupervised clustering approach discussed in section before known as the expectation maximization algorithm. The number of clusters was set to two for the failure and normal events and the resulting class distribution displayed a class imbalance problem with the normal class labels taking 87% of the data, and the failure class labels taking 13% of the data

Considering the class imbalance problem as discussed above, sampling techniques such as oversampling using SMOTE technique and under-sampling using the random sub-sample technique, were used to treat the imbalance presented in the dataset. Another imbalance technique considered was the cost-sensitive learning technique.

### 4.5 Training & Evaluation

Since the dataset containing over 870,000 instances is quite large, the dataset is split into training and test set using a 66% split. This means that the classifiers would be learning on about 574,200 instances and conducting classification on the remaining 295,800 instances. The split was selected to allow for the classifiers to learn more from the training set and have enough instances to conduct proper classification on. Cross Validation was considered for the evaluation of the dataset but because of hardware constraints which would lead to high computational complexity for the WEKA Suite, the training and test set split seemed a much better option

## 5. RESULTS

In this section, we present the results from the experiments conducted. The results would indicate which of the selected classifiers is/are the best performing algorithms to use for the creation of the ensemble model.

### 5.1 Experiments I

The following tables show the results of the performance measure of the classifiers when oversampling and undersampling were applied to the dataset. Also the tables show the performance measures after cost-sensitive learning using a cost matrix was applied to the dataset, and when no sampling or cost-sensitive learning was applied on the dataset.

**Table 1.** Performance comparison of ML techniques when no sampling has been applied to the dataset

| Measure | Machine Learning Techniques | | | | | | |
|---|---|---|---|---|---|---|---|
| | *J48* | *PART* | *MLP* | *NB* | *RF* | *LibSVM* | *SMO* |
| **TP Rate** | 0.983 | 0.983 | 0.981 | 0.979 | 0.982 | 0.971 | 0.980 |
| **FP Rate** | 0.090 | 0.086 | 0.057 | 0.032 | 0.081 | 0.028 | 0.059 |
| **Precision** | 0.983 | 0.983 | 0.981 | 0.981 | 0.982 | 0.975 | 0.981 |
| **Recall** | 0.983 | 0.983 | 0.981 | 0.979 | 0.982 | 0.971 | 0.981 |
| **F-Measure** | 0.983 | 0.983 | 0.981 | 0.980 | 0.982 | 0.972 | 0.981 |
| **ROC** | 0.990 | 0.998 | 0.991 | 0.998 | 0.998 | 0.972 | 0.961 |

**Table 2.** Performance comparison of ML techniques after oversampling has been applied to the dataset

| Measure | Machine Learning Techniques | | | | | | |
|---|---|---|---|---|---|---|---|
| | *J48* | *PART* | *MLP* | *NB* | *RF* | *LibSVM* | *SMO* |
| **TP Rate** | 0.982 | 0.982 | 0.981 | 0.978 | 0.983 | 0.969 | 0.980 |
| **FP Rate** | 0.039 | 0.031 | 0.022 | 0.012 | 0.034 | 0.011 | 0.014 |
| **Precision** | 0.982 | 0.982 | 0.982 | 0.980 | 0.983 | 0.973 | 0.981 |
| **Recall** | 0.982 | 0.982 | 0.981 | 0.978 | 0.983 | 0.969 | 0.980 |
| **F-Measure** | 0.982 | 0.982 | 0.981 | 0.979 | 0.983 | 0.970 | 0.981 |
| **ROC** | 0.994 | 0.995 | 0.991 | 0.999 | 0.999 | 0.979 | 0.961 |

**Table 3.** Performance comparison of ML techniques after undersampling has been applied to the dataset

| Measure | Machine Learning Techniques | | | | | | |
|---|---|---|---|---|---|---|---|
| | *J48* | *PART* | *MLP* | *NB* | *RF* | *LibSVM* | *SMO* |
| **TP Rate** | 0.981 | 0.981 | 0.980 | 0.978 | 0.981 | 0.971 | 0.978 |
| **FP Rate** | 0.019 | 0.019 | 0.020 | 0.021 | 0.019 | 0.029 | 0.021 |
| **Precision** | 0.981 | 0.981 | 0.980 | 0.979 | 0.981 | 0.972 | 0.979 |
| **Recall** | 0.981 | 0.981 | 0.980 | 0.978 | 0.981 | 0.971 | 0.978 |
| **F-Measure** | 0.981 | 0.981 | 0.980 | 0.978 | 0.981 | 0.971 | 0.978 |
| **ROC** | 0.989 | 0.988 | 0.993 | 0.998 | 0.998 | 0.971 | 0.990 |

**Table 4.** Performance comparison of ML techniques when using cost-sensitive learning with a cost matrix

| Measure | Machine Learning Techniques | | | | | | |
|---|---|---|---|---|---|---|---|
| | *J48* | *PART* | *MLP* | *NB* | *RF* | *LibSVM* | *SMO* |
| **TP Rate** | 0.983 | 0.983 | 0.981 | 0.981 | 0.983 | 0.958 | 0.975 |
| **FP Rate** | 0.111 | 0.113 | 0.071 | 0.072 | 0.102 | 0.180 | 0.115 |
| **Precision** | 0.983 | 0.983 | 0.981 | 0.981 | 0.983 | 0.957 | 0.975 |
| **Recall** | 0.983 | 0.983 | 0.981 | 0.981 | 0.983 | 0.958 | 0.975 |
| **F-Measure** | 0.982 | 0.982 | 0.981 | 0.981 | 0.982 | 0.957 | 0.975 |
| **ROC** | 0.991 | 0.982 | 0.991 | 0.998 | 0.998 | 0.889 | 0.991 |

## 5.2 Experiements II

This section of the paper describes the set of experiments conducted to compare the performance of the ensemble models created when trained on the dataset without sampling and cost-sensitive learning technique applied to the dataset (Table 5). The models created are:

- **Model I**: Stacked SMO using j48 & Multilayer perceptron
- **Model II**: Stacked SMO using RandomForest & Naïve Bayes
- **Model III**: Stacked SMO using PART, Multilayer perceptron and Naïve Bayes
- **Model IV**: Stacked SMO using RandomForest & PART
- **Model V**: Stacked SMO using RandomForest, Naïve Bayes and Multilayer perceptron.

Upon comparison of the best performing ensemble model, the model is then compared with the other classifiers on the dataset when no sampling technique or cost-sensitive learning has been applied (Table 6).

**Table 5.** Performance of ensemble models

| Measure | Ensemble Model | | | | |
|---|---|---|---|---|---|
| | *Model I* | *Model II* | *Model III* | *Model IV* | *Model V* |
| **TP Rate** | 0.983 | 0.982 | 0.983 | 0.983 | 0.982 |
| **FP Rate** | 0.112 | 0.081 | 0.086 | 0.114 | 0.085 |
| **Precision** | 0.983 | 0.982 | 0.983 | 0.983 | 0.982 |
| **Recall** | 0.983 | 0.982 | 0.983 | 0.983 | 0.982 |
| **F-Measure** | 0.982 | 0.982 | 0.983 | 0.982 | 0.982 |
| **ROC** | 0.936 | 0.998 | 0.998 | 0.934 | 0.993 |

**Table 6.** Performance of ensemble model compared with other Machine Learning techniques

| Measure | Machine Learning Techniques | | | | | | | |
|---|---|---|---|---|---|---|---|---|
| | *J48* | *PART* | *MLP* | *NB* | *RF* | *LibSVM* | *SMO* | *Model III* |
| **TP Rate** | 0.983 | 0.983 | 0.981 | 0.979 | 0.982 | 0.971 | 0.980 | 0.983 |
| **FP Rate** | 0.090 | 0.086 | 0.057 | 0.032 | 0.081 | 0.028 | 0.059 | 0.086 |
| **Precision** | 0.983 | 0.983 | 0.981 | 0.981 | 0.982 | 0.975 | 0.981 | 0.983 |
| **Recall** | 0.983 | 0.983 | 0.981 | 0.979 | 0.982 | 0.971 | 0.981 | 0.983 |
| **F-Measure** | 0.983 | 0.983 | 0.981 | 0.980 | 0.982 | 0.972 | 0.981 | 0.983 |
| **ROC** | 0.990 | 0.998 | 0.991 | 0.998 | 0.998 | 0.972 | 0.961 | 0.998 |

## 6. DISCUSSION

The results from the experiments conducted show the performance of some classifiers to be above the rest when sampling and cost sensitive learning techniques have been applied to the dataset. The PART and RandomForest classifier perform the best when initially trained on the raw dataset. With oversampling and under-sampling applied the Naïve Bayes and RandomForest technique are best performing with regards to the ROC measure, although the Naïve Bayes has a lower true positive rate (tpr) on both occasions. When applying cost-sensitive learning to the classifiers the true positive rate of the Naïve Bayes increases even as the ROC measure remains the same indicating that this classifier is a strong predictor. The same can also be said of the RandomForest classifier although this classifier has a tendency for overfitting the positive (normal) instances.

The prompt to compare different stacking models was brought about by the lack thereof of any significant distance between the performance measurements for the classifiers, apart from the LibSVM model which was the underperformer in all experiments conducted. From the table 5 above it can be seen that the two best performing models were model II and model III.

The better model was the model III as it correctly classified instances at 98.3018% compared to the 98.2312% from model II. The difference is not statistically significant because the model II has a lower false negative rate compared to model III.

The reasons for model III having more correctly classified instances is because of the extra base classifier, because the more prediction outputs for the meta-classifier to learn from, the better the decision making. This is made evident when the ensemble model is compared against the other classifiers as shown in table 6 when no sampling or cost-sensitive learning technique has been applied on the dataset. The model performs much better to the original SMO with a particular increase in the ROC value from 0.961 to 0.998. This increase can be attributed to the use of the stacking approach to improve the prediction capability of the SMO when learning from other classifiers.

## 7. CONCLUSION

In conclusion the review of literature has given insight into the notion of failure analysis and how it is being conducted by oil and gas personnel. Also the recent applications of failure analysis in various industries have been discussed. The methodology has reviewed the various approaches considered towards the creation of the ensemble model such as the expectation maximization clustering technique for assigning class labels, the sampling and cost-sensitive learning techniques for addressing the class imbalance, and the ensemble learning approach used for the creating the ensemble model.

The objectives of this research work have been achieved where it has been proved that the SMO technique can be used for analysis of oil and gas data than the libSVM technique. The created ensemble learning model using the stacking approach has also shown to be a better improved variant of the SMO technique. This clarifies the notion that ensemble models are better performers than original models on their own. The SMO training algorithm also shows it capability for use in failure analysis in the oil and gas industry.

Regarding any future implementation, although the classification has been addressed using the sample dataset obtained, the focus would be on applying this model to large oil and gas data, to determine if this model would be best used for analysis. This is because oil and gas data can be large in size, so a future approach would be to use a window based approach to take segments of the oil and gas dataset and then conducting prediction on it.

Another future implementation would be the use of the SMO for the creation of a pre-emptive incident engine, which can then be used to build various predictive models to make predictions on real-time oil and gas dataset. Also inference can also be drawn from the confusion matrix built from the predictive model as for how oil and gas experts can consider normal, failure events, and events which can lead to failure.

## 8. ACKNOWLEDGMENT

We would like to thank the Tentacle Technologies Sdn Bhd Big Data Analytics Lab for the assistance of conducting experiments.